# Creativity and Artificial Intelligence: A Digital Art Perspective

Bo Xing and Tshilidzi Marwala

University of Johannesburg

PO Box 524, Auckland Park, 2006

Republic of South Africa

bxing2009@gmail.com

**Abstract**

This paper describes the application of artificial intelligence (AI) to the creation of digital art. AI is a computational paradigm that codifies intelligence into machines. There are generally three types of AI and these are machine learning, evolutionary programming and soft computing. Machine learning is the statistical approach to building intelligent systems. Evolutionary programming is the use of natural evolutionary systems to design intelligent machines. Some of the evolutionary programming systems include genetic algorithm which is inspired by the principles of evolution and swarm optimization which is inspired by the swarming of birds, fish, ants etc. Soft computing includes techniques such as agent based modelling and fuzzy logic. Opportunities on the applications of these to digital art are explored.

## 1. Introduction

In the age of the 4$^{th}$ Industrial Revolution (4IR) (Xing and Marwala, 2017), many countries (Shah et al., 2015; Ding and Li, 2015) are setting out an overarching goal of building/securing an "innovation-driven" economy. As innovation emphasizes the implementation of ideas, creativity is typically regarded as the first stage of innovation in which generating ideas becomes the dominant focus (Tang and Werner, 2017; Amabile, 1996; Mumford and Gustafson, 1988; Rank et al., 2004; West, 2002). In other words, if creativity is absent, innovation could be just luck.

Though creativity can be generally understood as the capability of producing original and novel work or knowledge, the universal definition of creativity remains rather controversial, mainly due to its complex nature (Tang and Werner, 2017; Hernández-Romero, 2017). But putting it informally, by famous innovator Steve Jobs in 1995, we can think creativity like this way (Sanchez-Burks et al., 2015): "*Creative people [are] able to connect experiences they've had and synthesize new things.*" Artists are among the most creative people who take advantage of their senses all the time. By observing the surroundings, they continuously collect various raw material, which can be utilized later in their creative process.

On the other hand, one may notice the buzzword of artificial intelligence (AI) which, for now, can be broadly classified into three categories, namely, narrow AI, AGI (artificial general intelligence), and conscious AI (Wang and Goertzel, 2012). Currently active AI technologies mainly fall within the first group, i.e., narrow AI.

In this article, we attempt to use the digital art (Colson, 2007) – art in which digital technology stays in the center of its creative process – as an example of creativity to illustrate how creativity and AI can gain mutual benefit from each other.

## 2. Why Digital Art?

Technological development or revolution has a long history of influencing creativity. Take 'Gutenberg Revolution' (Kirschenbaum, 2016; Winston, 2005), the formation of the printing press in the fifteenth century paved the way to mass production of texts and images. With new communication capacity being enabled by this technological advancement, the widespread of material and intellectual exchange becomes possible. Nowadays, many of the working approaches used by digital artists can be traced back to the early days (between the 1950s and 1960s) of the computer development. Since the emergence of the World Wide Web in the 1990s, a diverse variety of opportunities were further opened up for visual arts with seemingly infinite permutable dimensions.

The reasons for choosing digital art as our discussion platform are threefold (Sefton-Green and Reiss, 1999; Bentkowska-Kafel et al., 2005). First, it is such a common practice for artists, in particular young professionals, to use a wide range of media arts for creative purposes, producing static/dynamic images, as well as manipulating sound tracks and text scripts. Second, digital art is not an isolated practice, divided from other forms of arts. It is essentially a methodology that incorporates all types of interconnections with other art exercises together with other manner of presentations and enquiries, illustrating that we are witnessing and experiencing a new wave of creativity revolution. Four, it is worth noticing that an army of digital artists are now working in numerous industries shoulder to shoulder with hardware and software practitioners at the forefront of innovation.

## 3. Creativity and Artificial Intelligence

AI techniques are at the forefront of creativity. They enable us to create huge new realms of innovative ways that otherwise would have never been imagined. Here, we sort them roughly into three movement-related groups, that is, short-term, mid-term, and long-term cases.

### *3.1 Short-Term Cases: AI-Assisted Digital Art Creation*

There are already many brave souls who have started a journey of finding out how creativity can be enhanced by AI. Among them, Matthew Yee-King, a musician who is employing evolutionary and genetic algorithm to work on sound synthesis. Meanwhile, Mario Klingemann, a code artist at the Google Cultural Institute, is applying deep learning to large image datasets for creating huge, stunning digital art. More details in this regard can be found from ArtFab at Carnegie Mellon University (Carnegie Mellon University, 2018).

### *3.2 Mid-Term Cases: Human and AI Collaborated Digital Art Creation*

Robots are marching towards us and the user friendly interface will become a crucial element for facilitating the interaction between human and robots (Xing and Marwala, 2018). Accordingly, the experience of taking the audience and artist relationship into account and forming a kind of new job like practical aesthetics is quickly blurring the conventional hard lines among algorithm programmer, engineering designer, and the artist. Other similar emerging new job titles also include Chief Experience Officer, Virtual Reality Editor, Mixed-

Reality Designer, Bot Developer, Hologram Retail Display Designer, and many more (Kapko, 2017).

*3.3 Long-Term Cases: Art Decentralized Autonomous Organization (ArtDAO)*

Nelson Mandela once said (BrainyQuote.com, n.d.): "*Money won't create success, the freedom to make it will.*" However, one of the main and long-standing disadvantages of the Internet is its hidden 'artist penalty', that is, creators find it hard to create and distribute digital content freely over the Internet while keeping themselves fairly remunerated. The ownership and copyright issue has thus become the biggest problems of digital art (Bentkowska-Kafel et al., 2005). With the progress of AI and some key infrastructural technologies, say, blockchain, this problem is more likely to be resolved after the appearance of Art Decentralized Autonomous Organization (ArtDAO), a preliminary version of Skynet in the market (Marwala and Hurwitz, 2017). In its simplest form, the ArtDAO works as follows (McConaghy and Holtzman, 2015): (1) Start and prepare yourself; (2) Generate new images by running AI art engine with the help of genetic programming or deep learning; (3) Announce the attribution via blockchain-bolstered platform, e.g., Ascribe; (4) Sell different editions on the marketplace, either centralized (e.g., Getty), or decentralized (e.g., OpenBazaar); (5) Claim remuneration from ArtDAO via cryptocurrency; (6) Check the termination criteria (e.g., whether you are satisfied with your income). If the stopping conditions are not met yet, then go to Step (7), that is, repeat the procedure and create more digital arts.

## 4. Outlook

Very few features of humanity are more charming than creativity; and rarely any other fields are more dynamic now than AI. The current usefulness of AI does not necessarily dwarf our human's creativity; on the contrary, the research and development of AI is largely spurred by our own internal creativity. Take interior search algorithm (Gandomi, 2014), the underlying inspiration sources were actually from the latest innovation in fine art (e.g., mirror work) and interior design process. Other examples also include music inspired algorithms, teaching–learning-based optimization, and so on (Xing and Gao, 2014b). A notable case goes to military operations inspired AI algorithms (Sun et al., 2016; Yang et al., 2018) which in turn can help us with modelling militarized conflict (Marwala and Lagazio, 2011). A bit outside AI realm, it was reported that the Tibetan knot (human artist's creation) has inspired German chemist Kekulé to finally discover the structure of benzene (Olteţeanu, 2016).

The domain of AI is rapidly evolving: narrow AI is good at pre-determined tasks; with AGI, we are targeting at solving any problem in general; while for conscious AI, a true or self-aware version of AI is envisioned. Poems were once regarded as a unique product of human creativity involving multiple dimensions of emotions and sentiments. However, during a recent Chinese national poem contest, an AI-powered robot beat three other rivals and won the champion prize (Lan, 2018). Does this mean that AI is able to 'think'? Is this dangerous or not? Interested readers can use (Lawless et al., 2017) as a reference for finding out more own interpretations.